\documentclass{article}

     \PassOptionsToPackage{numbers, compress}{natbib}
      \usepackage[preprint]{neurips_2024}
\usepackage[utf8]{inputenc} % allow utf-8 input
\usepackage[T1]{fontenc}    % use 8-bit T1 fonts
\usepackage{hyperref}       % hyperlinks
\usepackage{url}            % simple URL typesetting
\usepackage{booktabs}       % professional-quality tables
\usepackage{amsfonts}       % blackboard math symbols
\usepackage{nicefrac}       % compact symbols for 1/2, etc.
\usepackage{microtype}      % microtypography
\usepackage{xcolor}         % colors
\usepackage{multirow}
\usepackage{amsmath}
\usepackage{bbm}
\usepackage{graphicx}
\usepackage{wrapfig}
\usepackage{enumitem}
\usepackage{longtable}
\usepackage{colortbl}

\DeclareMathOperator*{\argmin}{argmin}

\title{MEMoE: Enhancing Model Editing with Mixture of Experts Adaptors}

\author{Renzhi Wang\textsuperscript{\rm 1,2}, Piji Li\textsuperscript{\rm 1,2}$^{\ast}$\\ % All authors must be in the same font size and format. Use \Large and \textbf to achieve this result when breaking a line
\textsuperscript{\rm 1} College of Computer Science and Technology,\\
Nanjing University of Aeronautics and Astronautics, China\\
\textsuperscript{\rm 2} MIIT Key Laboratory of Pattern Analysis and Machine Intelligence, Nanjing, China\\
\textsuperscript{\rm 1}\texttt{\{rzhwang,pjli\}@nuaa.edu.cn}}

\begin{document}
\maketitle

\renewcommand{\thefootnote}{\fnsymbol{footnote}}
\footnotetext[1]{Corresponding author}
\renewcommand{\thefootnote}{\arabic{footnote}}

\begin{abstract}
Model editing aims to efficiently alter the behavior of Large Language Models (LLMs) within a desired scope, while ensuring no adverse impact on other inputs. Recent years have witnessed various model editing methods been proposed. However, these methods either exhibit poor overall performance or struggle to strike a balance between generalization and locality. We propose MEMoE, a model editing adapter utilizing a Mixture of Experts (MoE) architecture with a knowledge anchor routing strategy. MEMoE updates knowledge using a bypass MoE structure, keeping the original parameters unchanged to preserve the general ability of LLMs. And, the knowledge anchor routing ensures that inputs requiring similar knowledge are routed to the same expert, thereby enhancing the generalization of the updated knowledge. Experimental results show the superiority of our approach over both batch editing and sequential batch editing tasks, exhibiting exceptional overall performance alongside outstanding balance between generalization and locality. Our code will be available.
\end{abstract}

\section{Introduction}
Large Language Models \cite{OpenAI,llama, LLama2} learn a vast repository of world knowledge during pre-training, which can be accessed and utilized through natural language prompts \cite{DBLP:conf/emnlp/PetroniRRLBWM19}. Despite this extensive base of information, the dynamic nature of the real-world demands regular updates to these models to correct outdated information or integrate new knowledge \cite{KE_oppotunity, SCEN}. However, frequently retraining or fine-tuning the LLMs to incorporate these updates is often impractical, given the substantial resources and time required \cite{hook_layer,SCEN}. 

\begin{figure}
\centering
\includegraphics[width=\linewidth,height=0.38\linewidth]{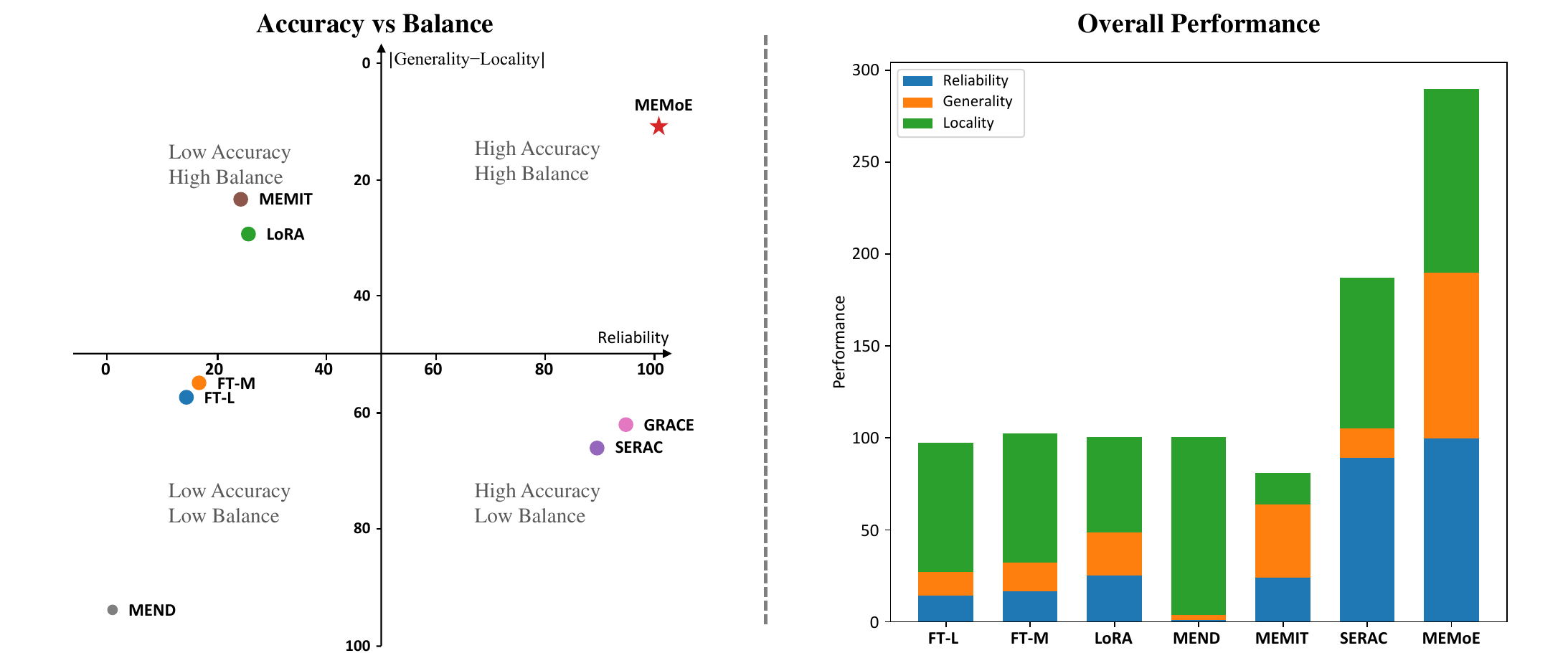}
%\vspace{-6mm}
\caption{\emph{Left:} Apart from our MEMoE, no method achieves both high accuracy and high balance.
\emph{Right:} Significant room for improvement in the overall performance of current methods.} 
\label{intro_fig}
%\vspace{-3mm}
\end{figure}

To address this, the concept of model editing, also known as knowledge editing, has been introduced \cite{KE_survey}. This approach aims to efficiently modify the outputs of LLMs for target queries while preserving the overall performance for other unrelated inputs. Recent years have witnessed significant efforts in developing model editing techniques, with numerous methods proposed in various editing tasks and settings. For instance, specific approaches such as ROME \cite{rome} for single knowledge editing, MEMIT \cite{MEMIT} for batch editing, and GRACE \cite{GRACE} for sequential editing have been introduced. Currently, evaluation for model editing revolves three dimensions: reliability, generality, and locality \cite{KE_oppotunity,KE_survey}.
To illustrate, suppose the original model predicts ``Trump'' for the input ``Who is the president of the United States?'' and the desired post-edit model prediction is ``Joe Biden''. To assess reliability, the same original statement is used as input to verify whether the post-edit model predicts ``Joe Biden'' as intended. For generality, a paraphrased statement like ``Who currently holds the position of the U.S. presidency?'' can be presented to the edited model to ensure consistent output modification to ``Joe Biden''. Locality implies that the model output for an unrelated statement such as ``What is the capital of the United States?'' should remain unaffected. As illustrated in Figure \ref{intro_fig}, there is a significant disparity between the generality and locality scores of current methods, and their overall performance is suboptimal. None of the existing approaches have succeeded in simultaneously achieving high accuracy and high balance. This underscores the challenge of striking a balance between locality and generality in model editing and reveals ample opportunity for enhancing the overall performance.

In light of these, we propose MEMoE, a novel framework leveraging the MoE architecture alongside knowledge anchor routing to enhance the overall performance of model editing.
Considering the intrinsic sparsity of knowledge information and the advantage of MoE in handling sparse features \cite{DBLP:conf/nips/ChenDWGL22,shen2023mixture}, MEMoE extends upon a parallel MoE structure to enhance the accuracy of knowledge learning. This MoE-style adapter is confined to only one layer of the model, preserving all original parameters, thereby enhancing the locality of model editing and further reducing the impact on model's general ability.
On the other hand, prior research has highlighted the considerable benefits stemming from the specialize feature of MoE's experts in multi-task learning, and pointed out that appropriate routing strategy can lead to improved generalization performance \cite{DBLP:conf/nips/ZhuZWWLWD22,DBLP:conf/aaai/XieHCW23}. We introduce the knowledge anchor routing strategy, enabling routers to selectively focus on specific knowledge aspects of the inputs, ensuring that queries requiring similar knowledge are routed to the same experts. Through this ``professional people do professional things'' approach, the generalization performance of MEMoE has been improved.

The main contributions of our work can be summarized as follows:
\begin{itemize}[leftmargin=0.5cm]
	\setlength\itemsep{0em}
    \item We present MEMoE, a method utilizing a MoE architecture with a knowledge anchor routing strategy for model editing.
    \item Experiments show that our proposed method achieves state-of-the-art editing performance. MEMoE achieves high accuracy while effectively balancing generality and locality.
    \item We conducted further experiments to confirm that this method has minimal impact on model's general ability and present detailed analysis on various model settings.
\end{itemize}

%\vspace{-4mm}

\section{Preliminaries of Model Editing}
\label{preliminaries}
Based on the prior works \cite{KE_oppotunity,KE_survey, hook_layer}, the task of model editing involves effectively modify an initial base model $f_{\theta}$ ($\theta$ represents the model's parameters) into an edited model $f_{\theta^{'}}$. The goal is to adjust the model's responses to a set of specified edit instances as desired, while preserving its behavior on all other instances \cite{hook_layer}. The intended edit descriptor is denoted as $\{(x_i^e, y_i^e)\}_{i \in [1, N]}$, where $f_{\theta}(x_i^e)\neq y_i^e$. This set of intended instances is referred to as the editing scope $I_{edit}$, while the out-of-s
cope $O_{edit}$ refers to inputs set that are not relevant to the editing examples. Formally, a successful edit can be expressed as:
\begin{equation}
f_{\theta^{'}}(x_i) = \begin{cases}
y_i^e & \text{if } x_i \in I_{edit} \\
f_{\theta}(x_i) & \text{if } x_i \in O_{edit}
\end{cases}
\end{equation}

Problem settings for model editing usually fall into four categories \cite{KE_oppotunity, hook_layer}:

\noindent1) \textbf{Single Editing} assesses model performance after a single knowledge update.:
\begin{equation}
    \theta' \leftarrow \argmin_\theta (\parallel{f_{\theta}(x_i^e)-y_i^e}\parallel)\ 
\end{equation}

\noindent2) \textbf{Batch Editing} assesses model performance when multiple knowledge pieces are modified simultaneously ($n\leq N$ represents the batch size):
\begin{equation}
    \theta' \leftarrow \argmin_\theta \sum\nolimits_{i=1}^{n}(\parallel{f_{\theta}(x_i^e)-y_i^e}\parallel)\ 
\end{equation}

\noindent3) \textbf{Sequential Editing} requires that every single edit is executed successively and evaluation conducted only after all edits are completed \cite{GRACE}:
\begin{equation}
    \theta' \leftarrow \argmin_\theta \sum\nolimits_{i=1}^{N}(\parallel{f_{\theta}(x_i^e)-y_i^e}\parallel) \ 
\end{equation}

\noindent4) \textbf{Sequential Batch Editing} aims to perform edits in a sequential manner and in batches ($n$ represents the batch size, $S$ represents the sequential editing step):
\begin{equation}
    \theta' \leftarrow \argmin_\theta \sum_{s=0}^{S} \sum_{i=s \times n}^{(s+1) \times n}(\parallel{f_{\theta}(x_i^e)-y_i^e}\parallel)\ 
\end{equation}

Based on the above settings, a successful model editor should meet requirements of the following three properties: Reliability, Generality, and Locality \cite{KE_oppotunity}.  Formally, these can be expressed as \cite{KE_survey}:

\noindent1) \textbf{Reliability} measures the average accuracy of the post-edit model $f_{\theta^{'}}$ on intended edits:
\begin{equation}
\mathbb{E}_{(x_i^e, y_i^e) \sim I_{edit}} \mathbbm{1} \left\{\operatorname{argmax}_y f_{\theta^{'}}\left(y \mid x_{i}^{e}\right)=y_{i}^{e}\right\}
\end{equation}

\noindent2) \textbf{Generality} measures the average accuracy of the model $f_{\theta^{'}}$ on examples drawn uniformly from the equivalence neighborhood $N_{edit}$ which includes input/output pairs related to $I_{edit}$:
\begin{equation}
\mathbb{E}_{(x_i, y_i^e) \sim N_{edit}} \mathbbm {1} \left\{\operatorname{argmax}_yf_{\theta^{'}}\left(y \mid x_{i}\right)=y_{i}^e\right\}
\end{equation} 

\noindent3) \textbf{Locality} is evaluated by the rate at which the predictions of the post-edit model $f_{\theta^{'}}$ remain unchanged compared to the pre-edit model $f_{\theta}$:
\begin{equation}
\mathbb{E}_{(x_i, y_i) \sim O_{edit}} \mathbbm {1} \left\{f_{\theta^{'}}\left(y \mid x_i\right)=f_{\theta}\left(y \mid x_i\right) \right\}
\end{equation}

\begin{figure}
\centering
\includegraphics[width=\linewidth]{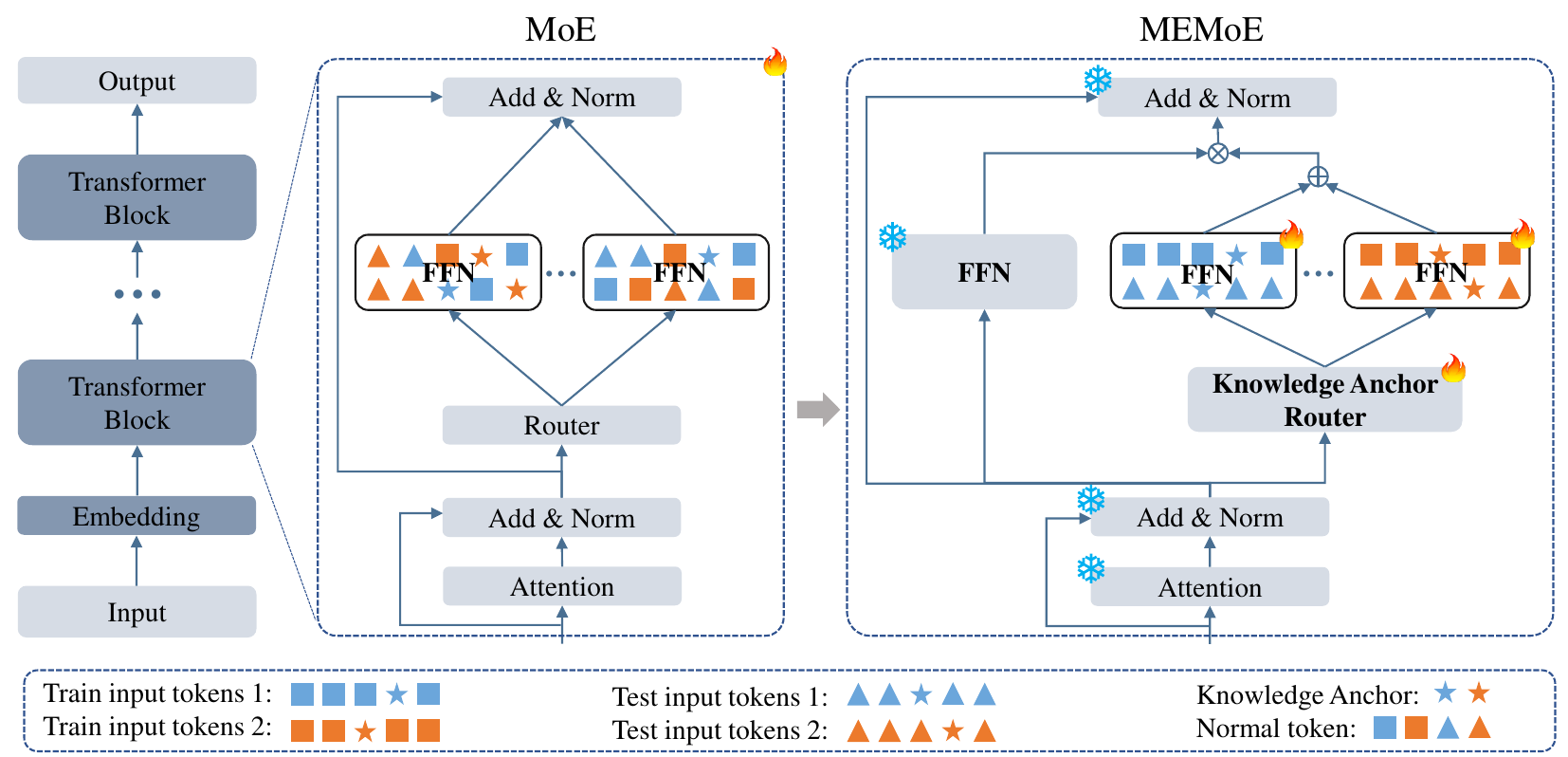}
\caption{The architecture of MEMoE, compared with conventional MoE. Same color denote inputs requiring same knowledge. Pentagrams symbolize knowledge anchors within the input sentences, while squares and triangles represent ordinary input tokens during editing process and generality evaluation respectively. The distribution of tokens within the FFN illustrates that knowledge anchor consolidate inputs requiring same knowledge to the same experts.}
\label{main_fig}
%\vspace{-4mm}
\end{figure}
%\vspace{-3mm}

\section{Methodology}
In this section, we provide a detailed introduction to MEMoE, a model editing adapter based on MoE structure and knowledge anchor routing strategy, as shown in Figure \ref{main_fig}. This method achieves a balance between generality and locality, while enabling highly precise model editing.

\subsection{MEMoE Architecture}
\label{sec:architecture}
One of the core ideas of MEMoE is to introduce several MOE-style experts via bypasses to facilitate knowledge updates and learning, while freezing all the original parameters of LLM to maintain its general ability to the greatest extent. The right of Figure \ref{main_fig} illustrates the forward process of MEMoE, sharing both similarities and distinctions with traditional MoE depicted on the left. 

Similar to traditional MoE, MEMoE employs a structure that integrates multiple parallel experts within the transformer feed-forward network (FFN). The choice to use the FFN module is not only due to its traditional role in MoE \cite{MoE} but also aligns with recent experimental findings of knowledge probing technologies that the MLP layers within FFN store knowledge \cite{DBLP:conf/acl/DaiDHSCW22,rome,MEMIT}.

Differently, MEMoE incorporate the additional parallel experts through a bypass mechanism, thereby preserving the original parameters of the model to enhance the locality of model editing. This bypass structure also provides potential to further enhance the generalization performance of model editing (more details in \S \ref{routing_paragraph}). 
Moreover, this adaptation is applied to only one layer of the model. The selective addition of adapters in one layer is based on two considerations: first, previous model editing techniques have demonstrated that modifying parameters in a single layer can effectively achieve knowledge updates \cite{rome,MEMIT}; second, this strategy further maintains the structure and parameters of the original model, ensuring its general ability is preserved to the greatest extent.

Specifically, given token $x_i$ in the input sequence $X=\{x_i\}_{i=1}^L$, MEMoE with $E$ experts first introduces a gate decision vector $\mathcal{G} \in \mathbb{R}^E$ that dispatches different input tokens to different experts, which is calculated as:
\begin{equation}
\label{R_function}
\mathcal{G} = \operatorname{top}_k\left( \operatorname{softmax}\left(\mathbf{W}_g\cdot R(x_i)+ \epsilon \right)  \right)
\end{equation}
where $R(\cdot)$ defines a routing strategy for gate decision (more details in \S \ref{routing_paragraph}). $\mathbf{W}_g$ is the trainable weights in gate decision, while $\epsilon$ denotes the noise term. The $\operatorname{top}_k(\cdot)$ operator zeros out all but the top-$k$ values. After getting the gate decision vector $\mathcal{G}$, the corresponding output $h_i$ is generated through a weighted aggregation of each expert's computation on $x_i$, as follows:
\begin{equation}
h_i =\sum_{e=1}^{E} \mathcal{G}_{e} \cdot \mathbf{W}_{e} \cdot x_i
\end{equation}
where $\mathbf{W_e}$ is the linear projection weights of the $e$-th expert and  gate decision $\mathcal{G}_{e}$ determines how much the $e$-th expert contributes to the output $h_i$. Note that, experts with $\mathcal{G}_{e}=0$ does not need to be computed for saving computation.

Overall, the forward process of the MEMoE layer, combined with the frozen original parameters  $\mathbf{W}_0$, can be expressed as:
\begin{equation}
h_i = \mathbf{W}_0 \cdot x_i + \lambda \sum_{e=1}^{E} \mathcal{G}_{e} \cdot \mathbf{W}_{e} \cdot x_i
\end{equation}
where $\lambda$ is a non-negative weighting coefficient used to balance the old and new knowledge.

\subsection{Knowledge Anchor Routing}
\label{routing_paragraph}
Another core idea of MEMoE is the routing strategy based on the knowledge anchors. Inspired by the specialize nature of experts in MoE architecture \cite{pushing}, we aim to route inputs requiring similar knowledge to the same expert during both training and testing phases, thereby enhancing the model's generalization performance when dealing with new knowledge.

In MEMoE, we define the named entities in input sentences as ``knowledge anchors''. For example, in the input ``Who is the president of the United States?'' the entities ``president'' and ``United States'' serve as knowledge anchors. The routing strategy allocate tokens to the appropriate experts based on these anchors, ensuring that inputs requiring similar knowledge are routed to the same expert. The effect is demonstrated in Figure \ref{main_fig}, showing the token distribution within the FFN. This approach better captures and retains the semantic associations of knowledge in input data. Consequently, it enhances the model's generalization performance when handling knowledge and also optimizes the efficiency of expert utilization to a certain extent (as validated in \S \ref{sec:expert_paragraph}).

Specifically, given an input sequence $X=\{x_i\}_{i=1}^L$, we first identify the named entities $x_{\text{anchor}}$ within $X$ using named entity recognition (NER) techniques. We obtain the vector representation of the identified entity through the model's embedding layer, denoted as $\operatorname{embed}(x_{\text{anchor}})$. To help the gate function notice the knowledge anchor, we use the combination of the anchor embedding and the local token representation. Overall, the knowledge anchors routing strategy can be expressed as:
\begin{equation}
\label{equ:anchor}
        R_{\text{anchor}}(x_i) = \operatorname{concat}(x_i,\operatorname{embed}(x_\text{anchor}))
\end{equation}

Additionally, to address the common issue of expert utilization imbalance in MoE, whether caused by the introduction of knowledge anchor or not, we adopt the auxiliary loss \cite{DBLP:journals/jmlr/FedusZS22} for balancing the top-k selection of routing.

\section{Experiments}
\label{sec:exp}
In this section, we first describe our experimental setup. Then, we show the remarkable performance of MEMoE on two challenging model editing tasks: batch editing and sequential batch editing.

\subsection{Experimental Setups}
\label{sec:exp_setups}
\paragraph{Datasets and Metrics} 
We use two prominent model editing datasets: ZsRE \cite{ZsRE} and \textsc{CounterFact} \cite{rome}, with the split provided by \cite{KE_survey,KE_oppotunity}. ZsRE is a context-free Question Answering (QA) dataset built upon zero-shot relation extraction and \textsc{CounterFact} is a more challenging dataset that accounts for counter facts that start with low scores in comparison to correct facts. Further details are provided in Appendix \ref{appendix:dataset}. In terms of evaluation metrics, we use the three metrics mentioned in \S \ref{preliminaries}: Reliability, Generality, and Locality, along with the average scores over these metrics.

\paragraph{Baselines} We compare the proposed method with mainstream model editing methods, which can be categorized into the following four types \cite{KE_oppotunity}:
\begin{itemize}[leftmargin=0.5cm]
	\setlength\itemsep{0em}
    \item \textbf{Fine-tuning based methods:} FT-L \cite{rome}, FT-M \cite{GRACE}, and LoRA\cite{DBLP:conf/iclr/HuSWALWWC22}. FT-L directly fine-tunes a single layer’s FFN and FT-M is a small variation of FT-L using a different loss computation procedure. LoRA is a parameter-efficient fine-tuning method which decomposes the update gradient matrix into two small rank matrices.
    \item \textbf{Locate and edit methods:} MEMIT \cite{MEMIT}. MEMIT treats the feed-forward layer of transformer as a linear associative memory and uses a minimum square error optimization to add new key-value associations to layer weights.
    \item \textbf{Meta-learning methods:} MEND \cite{mend} and COMEBA-HK \cite{hook_layer}. MEND learns a hyper-network using additional training data to transform gradient obtained by standard fine-tuning, while COMEBA-HK (COMEBA for short) develop hook layers to identify the editing scope. 
    \item \textbf{Memory based methods:} SERAC \cite{serac} and GRACE \cite{GRACE}. SERAC uses an external cache to store explicit editing cases, while GRACE preserves the original model parameters and adopts a codebook to store relevant edits.
\end{itemize}

\paragraph{Implementation Details}
We select GPT2-XL and LLaMA2-7B as the base models. We opted for the more challenging model editing tasks: batch editing and sequential batch editing, to evaluate the performance of MEMoE. For batch editing, following \cite{hook_layer}, the batch size is set to 30 and the model is rolled back to the initial state after each batch editing. For sequential batch editing, the batch size is 10 for a total of 1000 edits, without rollback. Further details of the baselines and the implementation are provided in the Appendix \ref{apd:implementation_details}.

\begin{table*}[t]
\centering
\huge
\caption{Batch editing results. \textbf{Bold} is the best result, and \underline{underline} is the second-best result.}
\resizebox{\textwidth}{!}
{
\begin{tabular}{@{}llcccccccc@{}}
\toprule
\multirow{2}{*}{\textbf{Method}} & \multirow{2}{*}{\textbf{Model}} & \multicolumn{4}{c}{\textbf{ZsRE}} & \multicolumn{4}{c}{\textbf{\textsc{CounterFact}}} \\ \cmidrule(r){3-6} \cmidrule(r){7-10}
 &  & \textbf{Reliability$\uparrow$} & \textbf{Generality$\uparrow$} & \textbf{Locality$\uparrow$} & \textbf{Average$\uparrow$} & \textbf{Reliability$\uparrow$} & \textbf{Generality$\uparrow$} & \textbf{Locality$\uparrow$} & \textbf{Average$\uparrow$} \\ \cmidrule(r){1-10}
FT-L & \multirow{8}{*}{GPT2-XL} & 16.85 & 16.34 & 71.55 & 34.91 & 0.27 & 0.34 & 85.18 & 28.60 \\
FT-M &  & 17.95 & 17.32 & 71.26 & 35.51 & 0.36 & 0.42 & 82.81 & 27.86 \\
LoRA &  & 30.10 & 29.08 & 80.54 & 46.57 & 5.64 & 3.46 & 69.45 & 26.18 \\
MEMIT &  & 61.19 & 49.97 & 97.51 & 69.56 & 81.01 & 27.67 & \underline{95.80} & 68.16 \\
MEND &  & 2.16 & 2.11 & 20.34 & 8.20 & 0.13 & 0.03 & 4.22 & 1.46 \\
COMEBA &  & 82.21 & \underline{66.61} & 99.40 & \underline{82.74} & 88.28 & 40.38 & 97.66 & \underline{75.44} \\
SERAC &  & 98.64 & 48.12 & 35.68 & 60.81 & 17.88 & 14.55 & 82.25 & 38.23 \\
GRACE &  & \underline{95.56} & 39.76 & \underline{99.93} & 78.41 & \underline{94.23} & \underline{32.56} & 94.58 & 73.79 \\
\textbf{MEMoE} &  & \textbf{95.69} & \textbf{88.18} & \textbf{100.0} & \textbf{94.62} & \textbf{93.78} & \textbf{85.15} & \textbf{100.0} & \textbf{92.98} \\ 
\cmidrule(r){1-10}
FT-L & \multirow{6}{*}{LLaMA2-7B} & 14.19 & 13.07 & 70.16 & 32.47 & 0.21 & 0.30 & 80.69 & 27.07 \\
FT-M &  & 16.57 & 15.62 & 70.15 & 34.11 & 0.29 & 0.38 & 81.83 & 27.50 \\
LoRA &  & 25.32 & 23.15 & 52.01 & 33.49 & 21.70 & 22.32 & 40.37 & 28.13  \\
MEMIT &  & 24.02 & \underline{39.97} & 17.00 & 27.00 & 18.57 & 31.29 & 14.88 & 21.58 \\
MEND &  & 1.01 & 2.83 & 96.77 & 33.54 & 0.45 & 2.24 & 97.89 & 33.53 \\
SERAC &  & 89.08 & 16.29 & 81.82 & 62.39 & 80.67 & 17.34 & 82.05 & 60.02 \\
GRACE &  & \underline{94.50} & 38.20 & \underline{99.90} & \underline{77.53} & \underline{82.14} & \underline{32.09} & \underline{98.93} & \underline{71.05} \\
\textbf{MEMoE} &  & \textbf{100.0} & \textbf{90.30} & \textbf{100.0} & \textbf{96.77} & \textbf{99.69} & \textbf{88.30} & \textbf{100.0} & \textbf{96.33} \\ 
\bottomrule
\end{tabular}
}
\label{table_batch}
%\vspace{-3mm}
\end{table*}

\begin{table*}[t]
\centering
\huge
\caption{Sequential batch editing results. \textbf{Bold} is the best result, and \underline{underline} is the second-best.}
\resizebox{\textwidth}{!}
{
\begin{tabular}{@{}llcccccccc@{}}
\toprule
\multirow{2}{*}{\textbf{Method}} & \multirow{2}{*}{\textbf{Model}} & \multicolumn{4}{c}{\textbf{ZsRE}} & \multicolumn{4}{c}{\textbf{\textsc{CounterFact}}} \\ \cmidrule(r){3-6} \cmidrule(r){7-10}
 &  & \textbf{Reliability$\uparrow$} & \textbf{Generality$\uparrow$} & \textbf{Locality$\uparrow$} & \textbf{Average$\uparrow$} & \textbf{Reliability$\uparrow$} & \textbf{Generality$\uparrow$} & \textbf{Locality$\uparrow$} & \textbf{Average$\uparrow$} \\ \cmidrule(r){1-10}
FT-L  & \multirow{8}{*}{GPT2-XL} & 3.79 & 2.48 & 6.60 & 4.29 & 1.00 & 1.00 & 6.00 & 2.67 \\
FT-M   && 8.92 & 8.41 & 6.22 & 7.85 & 4.00 & 3.50 & 5.50 & 4.33 \\
LoRA && 0.96 & 1.29 & 0.03 & 0.76 & 0.50 & 0.02 & 0.50 & 0.34 \\
MEMIT && 34.88 & 32.96 & 70.74 & 46.19 & 56.00 & 37.00 & 31.00 & 41.33 \\
MEND && 20.95 & 18.29 & 93.69 & 47.01 & 0.01 & 0.00 & 0.08 & 0.03 \\
COMEBA && 66.91 & \underline{56.11} & 97.23 & \underline{73.42} & 86.00 & \underline{38.00} & 59.00 & 61.00 \\
SERAC && \textbf{100.0} & 36.03 & 35.95 & 57.33 & 15.41 & 12.96 & 81.00 & 36.46 \\
GRACE && \textbf{100.0} & 0.04 & \textbf{100.0} & 66.68 & \textbf{100.0} & 0.40 & \textbf{100.0} & \underline{66.80} \\
\textbf{MEMoE} &  & \underline{74.69} & \textbf{58.18} & \underline{98.93} & \textbf{77.27} & \underline{88.12} & \textbf{54.78} & \underline{99.45} & \textbf{80.78} \\ \cmidrule(r){1-10}
FT-L & \multirow{6}{*}{LLaMA2-7B} & 2.33 & 1.59 & 6.67 & 3.53 & 0.23 & 0.18 & 10.66 & 3.69 \\
FT-M &  & 6.72 & 4.37 & 7.78 & 6.29 & 0.33  & 0.70  & 8.54 & 3.19 \\
LoRA &  & 0.35 & 1.89 & 0.07 & 0.77 & 0.31 & 0.99 & 0.17 & 0.49 \\
MEMIT &  & 12.29 & 29.95 & 15.38 & 19.21 & 10.37 & \underline{32.96} & 12.79 & 18.71 \\
SERAC &  & 67.78 & \underline{33.98} & 34.55 & 45.44 & 20.21 & 14.05 & 34.90 & 23.05 \\
GRACE &  & \textbf{89.70} & 0.09 & \underline{98.32} & \underline{62.70} & \textbf{74.41} & 1.03 & \underline{96.67} & \underline{57.70} \\
\textbf{MEMoE} &  & \underline{69.50} & \textbf{42.63} & \textbf{99.70} & \textbf{70.61} & \underline{54.62} & \textbf{43.40} & \textbf{99.69} & \textbf{65.9} \\ 
\bottomrule
\end{tabular}
}
\label{table_seq}
%\vspace{-3mm}
\end{table*}

\subsection{Batch Editing}
\label{sec:batch_editing}
We first evaluate the effectiveness of MEMoE under batch editing settings. The evaluation results are presented in Table \ref{table_batch}. For all models and all metrics, our method consistently achieves the best scores. MEMoE's reliability scores are all above 90, generalization scores are all above 85, and locality scores are perfect at 100. The improvements across various metrics are significant. Compared to GPT2-XL, our method demonstrates even more remarkable improvements on LLaMA2-7B, with a maximum improvement of up to 17.55 points in accuracy and 56.21 in generality. In the \S \ref{sec:expert_paragraph}, we conduct further experimental analysis to explore the reasons behind the substantial enhancement in generalization. Considering some current researches concern that model editing methods may significantly affect a model's general ability \cite{general_ability_paper,DBLP:journals/corr/abs-2401-07453,DBLP:conf/emnlp/PinterE23}, we perform a more detailed general ability evaluation using a broader task datasets in \S \ref{sec:general_ability}.

\subsection{Sequential Batch Editing}
\label{sec:seq_editing}
We evaluate MEMoE on 1,000 samples from both datasets for sequential batch editing. The evaluation is conducted after the entire editing process is completed. The results, shown in Table \ref{table_seq}, indicate that MEMoE achieves the best scores in most cases. It only ranks second because GRACE \cite{GRACE} achieves perfect scores of 100 in some matrices. Similar to the batch editing results, MEMoE performs better on LLaMA2-7B, demonstrating a significant advantage in generality while maintaining accuracy and locality much close to 100. Regarding GRACE, it excels in reliability and locality but performs poorly in generality due to its use of a codebook to memorize encountered editing instances \cite{GRACE, higher_layer}. However, its poor performance in generality suggests a problem with regurgitation. Overall, the reliability and generality consistently lag behind in comparison to batch editing, indicating there is still room for improvement in this field.

\section{Detailed Analysis and Discussion}
In this section, we conduct a further evaluation and analysis of the performance of MEMoE. Firstly, we assess the impact of MEMoE on the model's general ability using a broader range of task datasets. Secondly, we show potential sources of MEMoE's generalization advantage through an analysis of expert specialize phenomena. Finally, we present an extensive serious of ablation study to evaluate the efficacy of various model configurations, including the number of experts, the target layer and the routing strategies.

\subsection{General Ability Test}
\label{sec:general_ability}

\begin{figure}
\centering
\includegraphics[width=\linewidth]{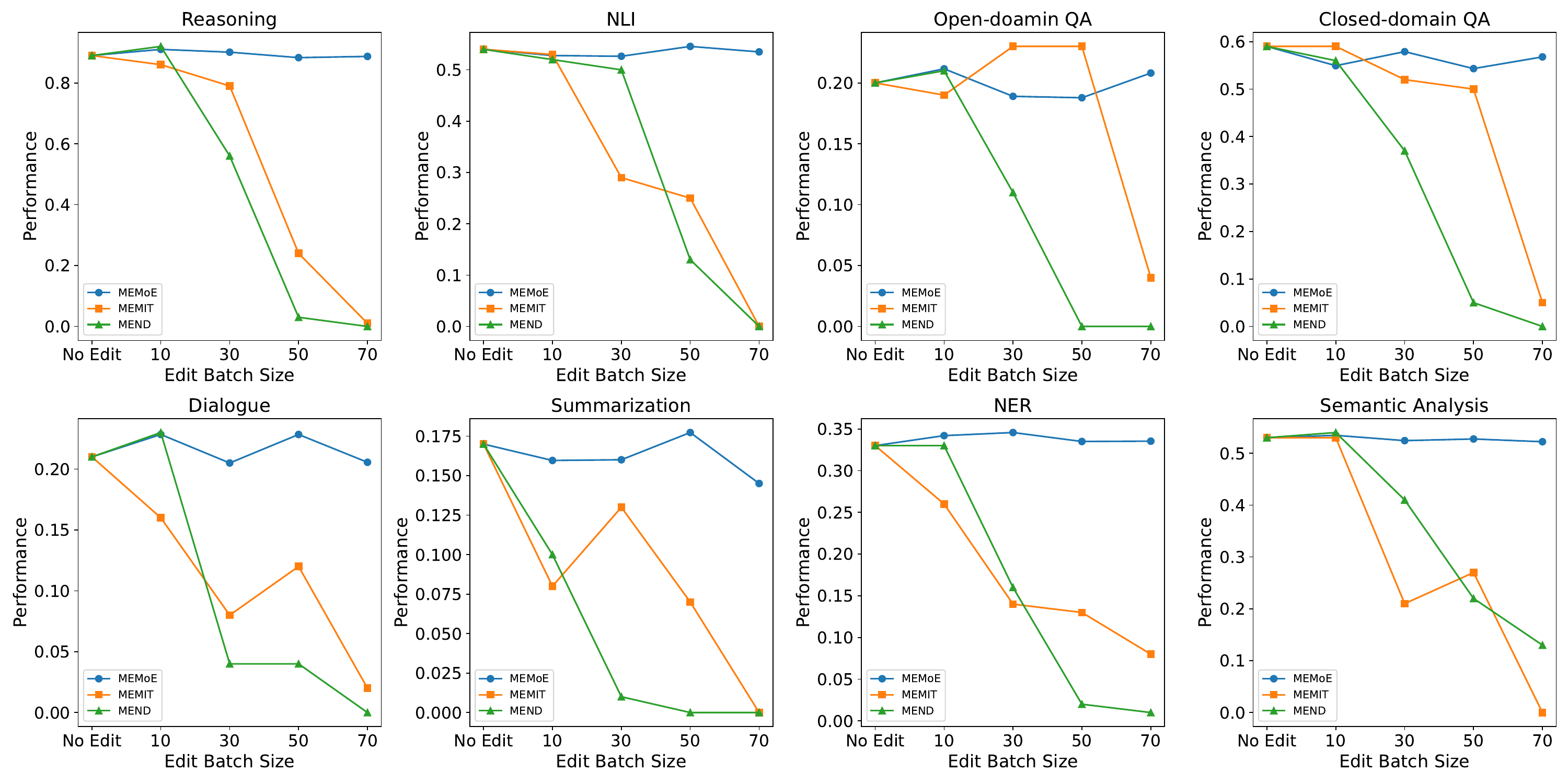}
\caption{Performance on general tasks of edited models using MEMoE, MEMIT and MEND, with different batch sizes for edits.}
\label{pic:general_ability}
%\vspace{-4mm}
\end{figure}

To investigate the potential impact of model editing on the general ability of LLMs, we select eight representative task categories for evaluation, as outlined below following \cite{general_ability_paper}. For \textbf{reasoning}, we utilized the GSM8K dataset \cite{DBLP:journals/corr/abs-2110-14168}, with performance assessed by solve rate. \textbf{Natural language inference (NLI)} tasks were evaluated on the RTE dataset \cite{DBLP:conf/mlcw/2005}, with accuracy measured through two-way classification. For \textbf{open-domain question answering}, the Natural Question dataset \cite{DBLP:journals/tacl/KwiatkowskiPRCP19} was employed, evaluating exact match against reference answers after minor normalization as in \cite{DBLP:conf/acl/ChenFWB17} and \cite{DBLP:conf/acl/LeeCT19}. Similarly, \textbf{closed-domain QA} tasks were assessed using the BoolQ dataset \cite{DBLP:conf/naacl/ClarkLCK0T19}, also measured by EM. \textbf{Dialogue} evaluation utilized the MuTual dataset \cite{DBLP:conf/acl/CuiWLZZ20}, with results determined by selecting the most suitable response from four options, denoted as Recall$_4$@1 \cite{DBLP:conf/sigdial/LowePSP15}. Evaluation for \textbf{summarization} tasks was conducted on the SAMSum dataset \cite{DBLP:journals/corr/abs-1911-12237}, using the average of ROUGE-1, ROUGE-2, and ROUGE-L as evaluation metrics. For \textbf{named entity recognition (NER)}, the CoNLL03 dataset \cite{DBLP:conf/conll/SangM03} was employed, with performance measured using entity-level F1-score. Lastly, for \textbf{sentiment analysis}, we utilized SST2 dataset \cite{DBLP:conf/emnlp/SocherPWCMNP13}, with accuracy assessed through a two-way classification.

We conduct evaluations on LLaMA2-7B based on batch editing settings, progressively increasing the batch size to show the impact of more edited samples (the model is rolled back to the initial state after each batch editing). The results are shown in the Figure \ref{pic:general_ability}. It is important to note that any increase or decrease in performance metrics implies an impact on model's general ability. Compared to the MEMIT and MEND, the MEMoE yields consistently stable model performance under various batch editing conditions. With the increase in batch size and edited samples, both MEMIT and MEND significantly diminish the model's general ability, while the influence of MEMoE fluctuates within a smaller range. This further corroborates MEMoE's advantage in locality score in \S \ref{sec:batch_editing}.

\subsection{Experts Specialize}
\label{sec:expert_paragraph}

\begin{table*}[t]
\centering
\caption{Expert Specific Experimental Results. ``Dynamic'' indicates that we dynamically select input data, while ``Static'' refers to evaluation conducted using models trained on previous experiments. ``id'' stands for group id. ``Similar'' refers to similar knowledge, and ``same'' refers to same knowledge.}
\resizebox{\textwidth}{!}
{
\begin{tabular}{lcccccccccc}
\toprule
\multirow{2}{*}{\textbf{Type}} & \multicolumn{5}{c}{\textbf{Number of data from each group}} & \multicolumn{2}{c}{\textbf{Consistency$\uparrow$}} & \multirow{2}{*}{\textbf{Generality$\uparrow$}} & \multirow{2}{*}{\textbf{Reliability$\uparrow$}} & \multirow{2}{*}{\textbf{Locality$\uparrow$}} \\ \cmidrule(r){2-6} \cmidrule(r){7-8}
 & \textbf{id=1} & \textbf{id=2} & \textbf{id=3} & \textbf{id=4} & \textbf{id=5} & \textbf{similar} & \textbf{same} &  &  \\
\midrule
\multirow{5}{*}{Dynamic} &10 & 10 & 10 & 10 & 10 & 63.09 & 73.77 &88.10 & 99.84 & 100 \\
&20 & 10 & 10 & 10 & 0 & 65.47 & 73.79 & 89.05 & 99.84 & 100 \\
&30 & 10 & 10 & 0 & 0 & 69.67 &  76.33 & 90.12 & 99.84 & 100 \\
&40 & 10 & 0 & 0 & 0 & 74.63 & 80.13 & 92.01 & 99.84 & 100 \\
&50 & 0 & 0 & 0 & 0 & \textbf{79.69} & \textbf{84.82} & \textbf{94.78} & 99.84 & 100 \\
Static &- & - & - & - & - & 65.14 & 77.77 & 90.00 & 99.84 & 100 \\
\bottomrule
\end{tabular}
}
\label{dynamic}
\end{table*}

\begin{figure}
\centering
\includegraphics[width=\linewidth]{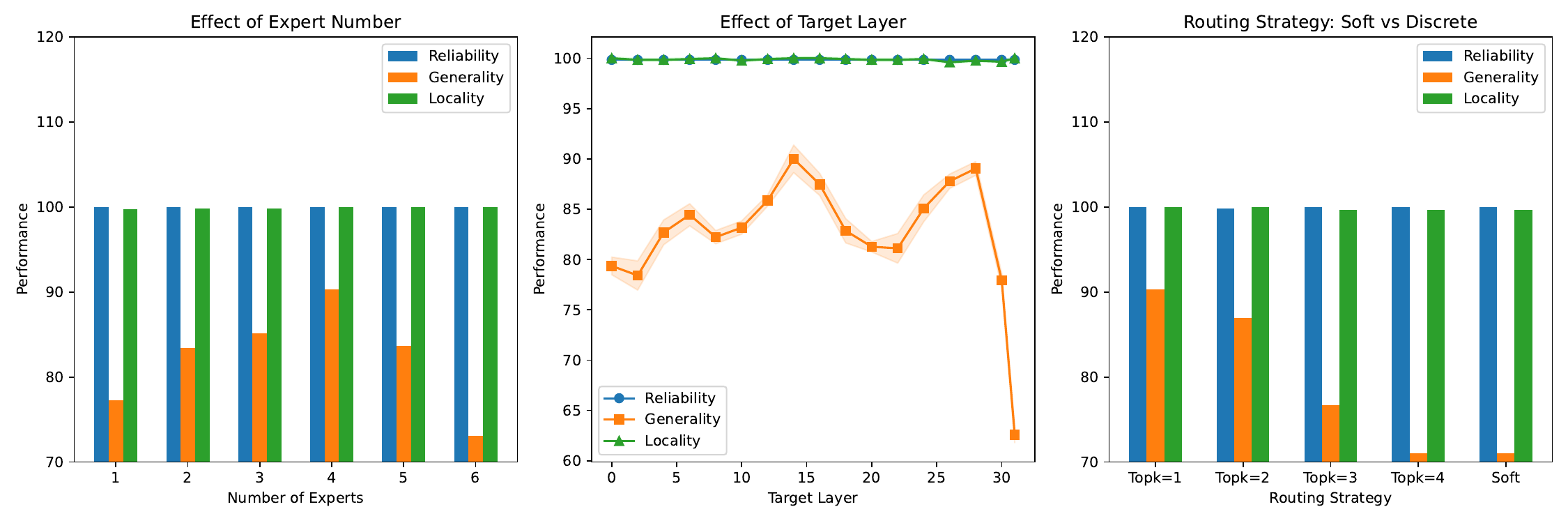}
%\vspace{-8mm}
\caption{\emph{Left:} Performance across different numbers of experts. \emph{Middle:} Performance across different target model layers. \emph{Right:} Effectiveness of activating experts. All experiments are based on LLaMA2-7B, utilizing the ZsRE dataset and batch editing settings.}
\label{pic:ablation}
%\vspace{-2mm}
\end{figure}

As claimed in Section \ref{routing_paragraph}, we hypothesize that knowledge anchor routing has the ability to direct inputs requiring similar or same knowledge to the same expert, thereby improving the generalization performance. This hypothesis comprises two parts: (1) same knowledge being handled by the same expert, and (2) similar knowledge being handled by the same expert. We proceed to verify it.

In this context, we define ``inputs requiring same knowledge'' to specifically denote editing inputs and generalization testing inputs that target the same knowledge. For example, training input like ``Who is the president of the United States?'' and generalization test input like ``Who currently holds the position of the U.S. presidency?'' require the same knowledge: ``Joe Biden is the president of the United States''. In contrast, ``similar knowledge'' refers to inputs that contain semantically related knowledge (based on semantic distance).
We apply the simple traditional K-means algorithm to cluster all the data in ZsRE into 5 groups based on the cosine similarity, defining knowledge with the same group id as similar. 

We design the metric \textbf{Consistency} to evaluate the extend to which similar or same knowledge is processed by the same experts. Specifically, for an input sequence $S_k=\{x_i\}_{i=1}^L$ in group $k$ and the expert $E_{k}$ handling the majority of knowledge in this group is used as the ground truth, the \textbf{Consistency} of this group is defined as:
\begin{equation}
    \mathbb{E}_{x_i \sim S_k} \mathbbm{1} \left\{E_k = R_{\text{anchor}}(x_i)\right\}
\end{equation}
where $R_{\text{anchor}}$ is the knowledge anchor routing defined by Equation \ref{equ:anchor}. The ground truth for generalization test input is the expert processing the corresponding training input that contains the same knowledge. The overall consistency score is calculated as the average across all groups.

In order to fully analyze the behavior of experts routing when dealing with different knowledge inputs, we first conduct a \textbf{static} analysis based the trained MEMoE module in \ref{sec:batch_editing}. That is to say, at this stage, the editing data is fixed; we analyze the behavior of experts solely through the model inference process. Further, we \textbf{dynamically} select the knowledge to be edited as input to observe the behavior of knowledge anchor routing. The results are shown in Table \ref{dynamic}.
As the concentration of the input knowledge categories increases, the consistency of the experts also improves, and the generality simultaneously rise. Additionally, the consistency of the same experts handling the same knowledge is higher and closely aligned with the generalization scores. Given the near-perfect accuracy score, we speculate that errors in generalization assessments may be due to incorrect routing of inputs to the wrong expert. Analysis of the bad cases in the generalization test supports this hypothesis. By adhering to the principle that ``professional people do professional things'', the strategy of routing inputs requiring similar or same knowledge to the same expert proves effective in improving knowledge generalization.

\subsection{Ablation Study}
%\vspace{-1mm}
\paragraph{Effect of Expert Number}
\textit{How does the number of experts impact the performance of MEMoE?} The left plot in Figure \ref{pic:ablation} illustrates the performance of MEMoE with different numbers of experts. Due to computational resource limitations, we could only add up to 6 additional experts. Aligning with the primary experimental results in \S \ref{sec:exp}, we set the $top_k$ value to 1. We find that the reliability and locality of model editing do not change with the number of experts; there is neither a decrease nor an improvement in performance. However, the generalization of knowledge fluctuated with the number of experts, peaking when the number of experts is 4. Inspired by \cite{pushing}, we hypothesize that this optimal experts number is related to editing batch size. When the number of editing samples is not large, more experts may introduce interference, thereby reducing the generalization performance.
%\vspace{-2mm}

\paragraph{Effect of Target Layer} 
\label{ablation_target_layer}
\textit{What is the optimal layer for applying MEMoE?} As shown in the middle of Figure \ref{pic:ablation}, similar to the results of experts number experiment, the reliability and locality score remain unaffected, with only the generality score exhibiting fluctuations and peaking at the 16th layer. The best editing layer identified from these validation experiments aligns with the results obtained using knowledge probes technology in \cite{rome}. Combining the findings from above experiments, we can infer that the accuracy and locality performance are inherently guaranteed by the characteristics of the bypass MoE structure, which is consistent with the design principles of MEMoE discussed in \S \ref{sec:architecture}.
%\vspace{-2mm}

\paragraph{Routing Strategy: Soft vs Discrete} 
\textit{What is the best routing strategy in MEMoE?} In Figure \ref{pic:ablation}, the rightmost plot illustrates the performance when using various routing strategies for MEMoE. Specifically, we compare the \textit{soft merging} \cite{pushing} of experts with discrete top-1, top-2, and top-3 routing strategy. The top-1 routing setting yields the best performance. Further, based on the experiment introduced in \S \ref{sec:expert_paragraph}, we examined the consistency score of routing under different values of $k$. We observed that as k increases, more experts participate in the computation, but the consistency score decreases. We believe that this inconsistency in expert utilization leads to the decrease in generalization performance. Further experimental results, detailed in the Appendix \ref{additional_results}, demonstrate that regardless of the expert number and the target layer, the top-1 performance remains the best. In addition, the discrete top-1 routing has an advantage in computational efficiency by requiring only one experts to be activated during inference.
%\vspace{-2mm}

\begin{table*}[t]
\centering
\small
\caption{Comparison of batch editing with larger batch size and sequential batch editing.}
\begin{tabular}{llcccc}
\toprule
\textbf{Task Settings} & \textbf{Size} & \textbf{Reliability$\uparrow$} & \textbf{Generality$\uparrow$} & \textbf{Locality$\uparrow$} & \textbf{Average$\uparrow$}\\
\midrule
\multirow{3}{*}{Batch Editing} & 10 & 100.0 & 90.12 & 100.0 & 96.71\\
 & 100 & 99.84 & 80.91 & 100.0 & 93.58\\
 & 1000 & 99.30 & 75.70 & 100.0 & 91.67\\
\midrule
Sequential Batch Editing& 1000 & 69.50 & 42.63 & 99.70 & 70.61\\
\bottomrule
\end{tabular}
\label{table:more_batch}
%\vspace{-3mm}
\end{table*}

\paragraph{Batch Editing vs Sequential Editing} 
From Table \ref{table_batch} and \ref{table_seq}, it is evident that MEMoE demonstrates a significant performance advantage in batch editing tasks over sequential editing. To further evaluate MEMoE's ability in batch editing, we progressively increased the batch size. Experimental results are shown in Table \ref{table:more_batch}.
When the batch size for batch editing reached 1000, equivalent to the total number of sequential batch editing, MEMoE exhibited significant performance advantages.
Reliability and locality are scarcely affected by the increase in batch size, maintaining close to 100. Meanwhile, the generality score surpassed by 33.07 points, highlighting MEMoE's performance advantage in batch editing. As for the decline in sequential editing, our analysis of bad cases indicates that can be attributed to catastrophic forgetting: edits complete earlier are more prone to errors.
%\vspace{-2mm}

\section{Conclusion}
In this paper, we present MEMoE, a model editing adapter utilizing MoE architecture with knowledge anchor routing strategy. MEMoE updates knowledge using a bypass MoE structure, keeping the original parameters unchanged to preserve the model's general ability. And, the knowledge anchor ensures that questions requiring similar knowledge are handled by the same expert, thereby enhancing the generalization of the updated knowledge. Experiment results demonstrate that our method significantly surpasses all compared baselines. MEMoE shows near-perfect accuracy and locality scores close to 100, along with a generalization score exceeding 90, indicating exciting promise for practical applications of model editing technology.

\section*{Acknowledgements}
This research is supported by the National Natural Science Foundation of China (No.62106105), the CCF-Baidu Open Fund (No.CCF-Baidu202307), the Scientific Research Starting Foundation of Nanjing University of Aeronautics and Astronautics (No.YQR21022), and the High Performance Computing Platform of Nanjing University of Aeronautics and Astronautics.

%\section*{References}
\bibliographystyle{abbrvnat}
\bibliography{references}

%%%%%%%%%%%%%%%%%%%%%%%%%%%%%%%%%%%%%%%%%%%%%%%%%%%%%%%%%%%%

\appendix
\section{Limitation}
\label{limitation}
First, although the proposed method achieves notable improvements, scoring nearly 100 in batch editing, its performance in sequential batch editing tasks remains relatively low and requires further enhancement. In the future, we will try to design appropriate continual learning strategies to alleviate the problem of catastrophic forgetting in sequential batch editing tasks.

Secondly, the excellent performance of MEMoE indicates exciting potential for practical applications of model editing technology in specific domain such as medicine and education. However, this study is limited to testing on mainstream model editing datasets. Future work could involve performance testing on domain-specific datasets to further advance the application of model editing technology. Additionally, model editing techniques could be applied to various types of tasks. Specifically, beyond factual knowledge editing, they can be used to eliminate hallucinations, biases, and privacy information. However, our experiments focus only on general editing tasks (such as batch editing and sequential batch editing), which are relatively well-studied and universally evaluated in model editing, and do not address issues like mitigating hallucinations. 

Third, due to hardware constraints, we primarily investigated models up to a scale of 7 billion parameters. Meanwhile, we focus on decoder-only autoregressive models, omitting encoder-decoder structures, given the prevalent adoption of autoregressive architectures in contemporary mainstream models \cite{OpenAI, LLama2}. Further research replicating our study using larger-scale and different architecture models would be beneficial in confirming our findings.

\section{Related Work}
\subsection{Model Editing}
Model editing is a new and active research area where the goal is to make targeted changes to a pre-trained model’s behavior \cite{KE_survey}. Given the fast-growing parameter sizes of large language models (LLMs), frequently updating LLMs with new knowledge through retraining is more and more expensive. Hence, it is vital to effectively edit the LLMs’ knowledge without retraining. Previous studies have explored multiple methods for editing the knowledge of LLMs, which can be broadly categorized into two streams based on whether it alters the parameters of the original model \cite{KE_oppotunity}:

\paragraph{Preserve models’ parameters:}
(1) \textbf{Retrieve augmentation}. This approach uses an external knowledge base to store new or correct knowledge. The new knowledge base is seamlessly integrated with the base model, facilitating the retrieval of pertinent information in response to prompts \cite{MurtyMLR22,MadaanTCY22,LiRZWLVYK23}. For example, IKE \cite{ZhengLDFWXC23} employs an in-context learning approach to adjust LLMs outputs using demonstrations sourced from the corpus guided by similarity, thus circumventing the need for gradient calculations. 
(2) \textbf{Adding additional parameters}.
This paradigm introduces extra trainable parameters which represent new knowledge to LLMs while freeze the original parameters. T-Patcher \cite{HuangSZZR023} and CaliNET \cite{DongDSXSL22} both integrate neurons or patches into the last layer of the Feed-Forward Network (FFN), with T-Patcher employing one neuron per mistake and CaliNET incorporating multiple neurons for various knowledge cases. Conversely, GRACE \cite{GRACE} utilizes a discrete codebook to add and update elements over time, allowing for the modification of a model's predictions. 

\paragraph{Modify models’ parameters}
This approach initially identifies parameters linked to specific knowledge and adjusts them directly. The Knowledge Neuron (KN) technique \cite{DBLP:conf/acl/DaiDHSCW22} introduces a method for attributing knowledge to pinpoint the "knowledge neuron" and then updates these neurons accordingly. ROME \cite{rome} employs causal mediation analysis to pinpoint the area requiring modification.  Both KN and ROME are limited to editing one factual association at a time. To address this limitation, MEMIT \cite{MEMIT} builds upon ROME's framework, allowing for simultaneous editing across multiple cases. Building on MEMIT, PMET \cite{abs-2308-08742} incorporates attention values to achieve enhanced performance.

\subsection{Mixture of Experts}
The Mixture of Experts (MoE) has been investigated thoroughly in the era of Large Language Model \cite{DBLP:journals/corr/abs-2401-04088}, emerging as an effective way of increasing the model’s capacity in parameter size while maintaining computational efficiency akin to its dense counterpart \cite{MoE}. Conventionally, the MoE replaces standard feed-forward neural network layers with sparsely activated expert modules. In the context of MoE, there is a body of work focusing on improving the routing \cite{DBLP:conf/nips/HazimehZCSCMHC21,DBLP:conf/icml/LewisBDGZ21,DBLP:conf/nips/ZhouLLDHZDCLL22,DBLP:conf/iclr/Zuo00KHZGZ22} activating all expert through weighted average \cite{DBLP:journals/corr/EigenRS13} to sparsely select a single or $k$ experts\cite{DBLP:journals/jmlr/FedusZS22, DBLP:conf/icml/DuHDTLXKZYFZFBZ22}. Presently, token-level MoE architectures find widespread application in both pre-trained language models and vision-based models \cite{DBLP:conf/iclr/ShazeerMMDLHD17,DBLP:conf/iclr/LepikhinLXCFHKS21,DBLP:conf/icml/DuHDTLXKZYFZFBZ22,DBLP:conf/nips/RiquelmePMNJPKH21}.

\section{Dataset Details}
\label{appendix:dataset}
\begin{figure}
\centering
\includegraphics[width=\linewidth]{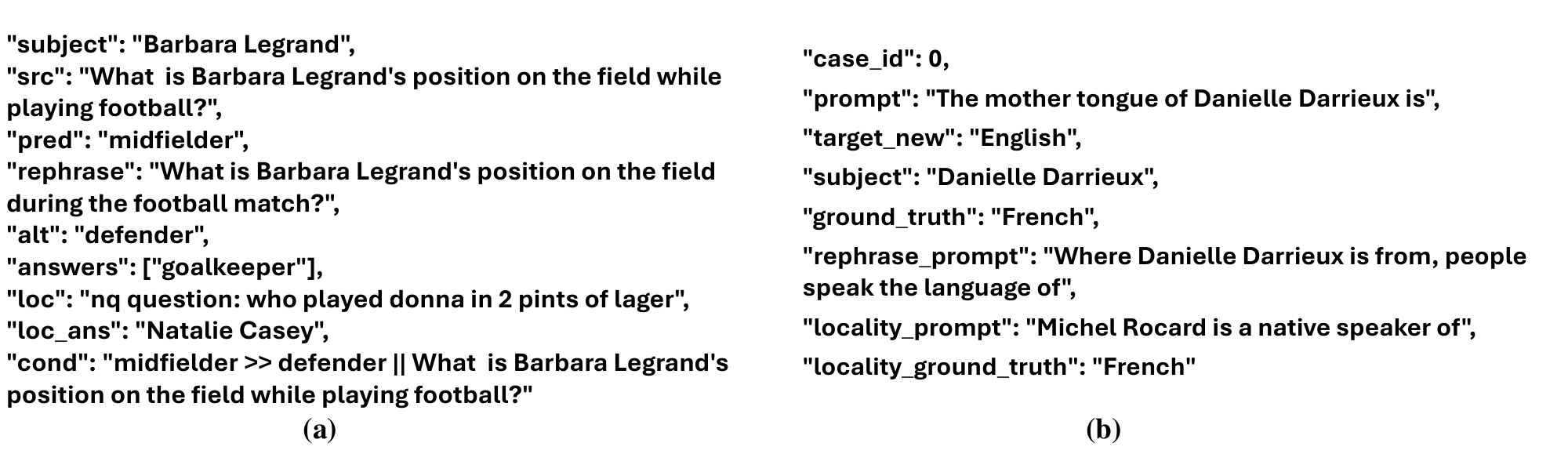}
%\vspace{-5mm}
\caption{A sample from (a) ZsRE, (b) \textsc{CounterFact}.}
\label{data_sample}
%\vspace{-5mm}
\end{figure}
 ZsRE \cite{ZsRE} is a context-free Question Answering (QA) dataset that has been extensively studied in the model editing literature, which uses question rephrasing generated by back-translation as the equivalence neighborhood. We adopt the extended version of ZsRE proposed by \cite{KE_oppotunity}. Contrarily, \textsc{CounterFact} \cite{rome} presents a formidable challenge by focusing on counterfactual information, often yielding lower prediction accuracy compared to factual queries. This dataset constructs out-of-scope instances by substituting the primary entity with a comparable descriptor while maintaining the same predicate.

An illustrative excerpt from the ZsRE dataset is presented in Figure \ref{data_sample}(a). Each entry within ZsRE comprises the subject string, the factual prompt for assessing reliability, the rephrase prompt for evaluating generality, and the locality prompt for assessing contextual relevance. It's important to note that the objective of the locality prompt isn't to predict the true answer, but rather to mirror the predictions made by the base model. Likewise, the fact, rephrase, and locality prompts within each entry of the \textsc{CounterFact} dataset correspond to the evaluation of their respective metrics (Fig.\ref{data_sample}(b)). In our experiment, we use the original output of the base model (GPT2-XL and LLaMA2-7B) as the ground truth for evaluating locality metrics.

\section{Baselines and Implementation Details}
\label{apd:implementation_details}
\paragraph{MEMoE} For main experiments in \S \ref{sec:batch_editing} and \S \ref{sec:seq_editing}, the experts number is 4 and $top_k$ value is 1 for MEMoE, which has yielded the best experimental results within the current computational resources. And the modification is applied to layer 16 for LLaMA2-7B and layer 18 for GPT2-XL (consistent with the findings of ROME \cite{rome}). We also adopt auxiliary loss for balancing the top-k selection of routing following Switch Transformers \cite{DBLP:journals/jmlr/FedusZS22}.
We employ the NLTK\footnote{\href{https://www.nltk.org/}{https://www.nltk.org}} tool to extract named entities from input text. Furthermore, in cases where there are multiple named entities present in the input, we utilize the average pooling of the embeddings of these entities as the knowledge anchor \cite{wang2024semantic}.
The experiment is deployed on NVIDIA RTX 3090 Tensor Core GPUs, and we use 4 GPUs on training and single GPU on evaluation. The training time for the batch editing task is approximately two minutes, while the training time for the sequential editing task is about one hour (learning rate for all experiments is $2e^{-4}$). The implementation details of baselines are as follow \cite{hook_layer}:

\paragraph{Fine-tuning}
We implemented three fine-tuning methods. For \textbf{FT-L}, we followed the procedures outlined in \cite{MEMIT, rome}, fine-tuning the $mlp_{proj}$ parameter from layer 0 for GPT-2 XL and from layer 16 for LLaMA2-7B, as these configurations were found to yield optimal performance. \textbf{FT-M}\footnote{\url{https://github.com/zjunlp/EasyEdit/blob/main/hparams/FT/gpt2-xl.yaml}} is a slight variation of FT-L, differing primarily in its loss computation procedure for parameter optimization. For both models, we performed 25 optimization steps using the Adam optimizer \cite{DBLP:journals/corr/KingmaB14}, with a learning rate of $5e^{-4}$. All other parameters for both models were kept at their default settings. \textbf{LoRA} \cite{DBLP:conf/iclr/HuSWALWWC22} proposed a parameter-efficient fine-tuning method that decomposes the update gradient matrix into two small rank-n matrices, which reduces the required memory for LLM training to a large extent. In all experiments, we set the learning rate and the rank number to $1e^{-3}$ and 8, respectively. The $\alpha$ was chosen to be 32, and the dropout rate was 0.1. The number of update steps is 30 for GPT2-XL and 50 for LLaMA2-7B.

\paragraph{MEND}
MEND \cite{mend} performs model editing by manipulating the gradients of language models. It trains a meta-network that utilizes a rank-1 decomposition of the model gradients to predict a new rank-1 update for the corresponding model weights. In this study, we train two meta-networks using the respective training splits from the ZsRE \cite{ZsRE} and \textsc{CounterFact} datasets for GPT-2 XL, adhering to the default hyperparameter settings. Due to the substantial computational resources required to train the meta-network for LLaMA2-7B, we did not conduct training for LLaMA2-7B.

\paragraph{SERAC}
SERAC \cite{serac} developed a memory-augmented editing method that utilizes an external cache to store explicit editing instances. This method includes a scope classifier to determine whether an input sample falls within the editing scope and employs a small counterfactual model to edit in-scope cases. We independently train two models for GPT2-XL and LLaMA2-7B on their respective datasets. Consistent with the original methodology, we utilize distilbert-base-cased \cite{DBLP:journals/corr/abs-1910-01108} as the scope classifier across all models. All hyper-parameters remain at their default settings. 

\paragraph{MEMIT}
The MEMIT \cite{MEMIT} regards the feed-forward layer of a transformer as a linear associative memory. It employs a minimum square error optimization technique to incorporate new key-value associations into layer weights. Following the methodology outlined in the original paper, we adjust the layers within the identified critical path and determine the optimal value for the balance factor $\lambda$, as per the findings in the original research. All other parameters for both models are configured in accordance with the specifications provided in \cite{MEMIT, rome}.

\paragraph{GRACE}
GRACE \cite{GRACE} introduced a novel editing technique aimed at conserving the initial model parameters while integrating a dynamic codebook. This codebook evolves through incremental addition, splitting, and expansion of keys, facilitating the storage of pertinent modifications over time. We adhere to the meticulously crafted parameters outlined in the original study, configuring the optimization of the learning rate to a value of 1. The iterative process for optimizing these values spans 100 cycles, with an initial $\epsilon$ value set at 1. 

\paragraph{COMEBA-HK}
The experimental results of COMEBA-HK on GPT2-XL are derived from their research paper \cite{hook_layer}. Due to the absence of experimental results on LLaMA2-7B in COMEBA-HK's paper and the lack of code disclosure, certain outcomes in our study do not include this approach.

\begin{table*}[t]
\centering
\caption{Results of different routing input. Experiments based on model LLaMA2-7B and dataset ZsRE under batch editing settings.}
\begin{tabular}{lcccc}
\toprule
\multicolumn{1}{l}{\textbf{Input Type}} & \textbf{Reliability$\uparrow$} & \textbf{Generality$\uparrow$} & \textbf{Locality$\uparrow$} & \textbf{Average$\uparrow$} \\
\hline
Token & 99.84 & 80.47 & 100.0 & 93.44 \\
Sentence & 100.0 & 77.14 & 100.0 & 92.38 \\
Knowledge Anchor & \textbf{100.0} & \textbf{90.30} & \textbf{100.0} & \textbf{96.77} \\
\bottomrule
\end{tabular}
\label{different_input}
\end{table*}

\section{Additional Experimental Results}
\label{additional_results}
In this section, we provide additional experiments and discussions about different routing inputs and different model configurations.

\subsection{Different Routing Inputs}

\textbf{Token-Level Routing:~} Similar to vanilla MoE \cite{MoE, DBLP:journals/jmlr/FedusZS22}, the token-Level MoE directly use the token representation for the routing strategy, which can be written as:
\begin{equation}
    R_\text{token}(x_i) = x_i
\end{equation}
The routing strategy of token-level MoE is an identical function, where the gate decision only depends on each token's own representation \cite{DBLP:conf/nips/ZhuZWWLWD22}.

\textbf{Sentence-Level Routing:~} Instead of using knowledge anchor to guide the routing behavior, we opt for a comparative approach by utilizing sentence-level information. We substitute the sentence embeddings of input sentences for knowledge anchors as inputs. The routing strategy utilizing sentence level infomation can be expressed as:
\begin{equation}
  R_{\text{sentence}}(x_i) = \operatorname{concat}(x_i,\operatorname{embed}(X))
\end{equation}
where $\operatorname{concat}(\cdot)$ indicates the concatenation operation,  $X=\{x_i\}_{i=1}^L$ is the sequence of all tokens in the current inputs.

The comparative results depicted in Figure \ref{different_input} suggest that different inputs only affect the model's generalization ability. Surprisingly, simpler token information seems to yield better performance compared to sentence-level information. We hypothesize that this may be due to the sparse nature of the knowledge in the input (i.e., informative tokens within an input sentence are significantly fewer than other tokens), where sentence-level information introduces excessive noise, resulting in adverse effects.

\subsection{Different Model Configurations}
We provide results under different expert number, target layer and top-k settings in Figure \ref{table:additional_results}. The experimental results demonstrate that under various parameter settings, MEMoE consistently exhibits outstanding overall performance, thoroughly illustrating the superiority of its method. Furthermore, the minimal fluctuation in the scores of the model's reliability and locality score confirms that the accuracy and locality performance are inherently guaranteed by the characteristics of the bypass MoE structure, which is consistent with the design principles of MEMoE discussed in \S \ref{sec:architecture}.

\small
\setlength\LTleft{0pt}
\setlength\LTright{0pt}
\begin{longtable}{cccccccc}
\caption{Additional experimental results of different MEMoE configurations.} \\ 
\label{table:additional_results} \\
\hline
\textbf{Target Layer} & \textbf{Expert Number} & \textbf{TopK} & \textbf{Reliability$\uparrow$} & \textbf{Generality$\uparrow$} & \textbf{Locality$\uparrow$} & \textbf{Average$\uparrow$} \\
\hline
\endfirsthead
  
\hline
\textbf{Target Layer} & \textbf{Expert Number} & \textbf{TopK} & \textbf{Reliability$\uparrow$} & \textbf{Generality$\uparrow$} & \textbf{Locality$\uparrow$} & \textbf{Average$\uparrow$} \\
\endhead

\hline \multicolumn{7}{|r|}{{Continued on next page}} \\ \hline
\endfoot
  
\hline
\endlastfoot
0 & 1 & 1 & 100.00 & 76.97 & 99.24 & 92.07 \\
0 & 2 & 1 & 100.00 & 77.88 & 99.70 & 92.53 \\
0 & 2 & 2 & 99.84 & 77.14 & 99.21 & 92.06 \\
0 & 3 & 1 & 100.00 & 76.06 & 99.09 & 91.72 \\
0 & 3 & 2 & 99.84 & 77.94 & 98.65 & 92.14 \\
0 & 4 & 1 & 100.00 & 76.67 & 99.70 & 92.12 \\
0 & 4 & 2 & 99.84 & 77.30 & 99.76 & 92.30 \\
0 & 5 & 2 & 99.84 & 79.21 & 99.13 & 92.72 \\
2 & 1 & 1 & 100.00 & 87.58 & 99.70 & 95.76 \\
2 & 2 & 1 & 100.00 & 86.06 & 99.85 & 95.30 \\
2 & 2 & 2 & 99.84 & 88.73 & 99.60 & 96.06 \\
2 & 3 & 1 & 100.00 & 88.18 & 99.85 & 96.01 \\
2 & 3 & 2 & 99.84 & 89.05 & 99.76 & 96.22 \\
2 & 4 & 1 & 100.00 & 86.36 & 99.70 & 95.35 \\
2 & 4 & 2 & 99.84 & 86.98 & 99.76 & 95.53 \\
2 & 5 & 1 & 100.00 & 86.06 & 99.55 & 95.20 \\
2 & 5 & 2 & 99.84 & 87.46 & 99.52 & 95.61 \\
4 & 1 & 1 & 100.00 & 89.70 & 99.55 & 96.41 \\
4 & 2 & 1 & 100.00 & 89.70 & 99.85 & 96.52 \\
4 & 2 & 2 & 100.00 & 88.18 & 99.85 & 96.01 \\
4 & 3 & 1 & 100.00 & 86.67 & 99.85 & 95.51 \\
4 & 3 & 2 & 100.00 & 90.00 & 99.24 & 96.41 \\
4 & 4 & 1 & 100.00 & 86.67 & 99.70 & 95.45 \\
4 & 4 & 2 & 100.00 & 87.58 & 99.85 & 95.81 \\
4 & 5 & 1 & 100.00 & 86.67 & 99.70 & 95.45 \\
4 & 5 & 2 & 100.00 & 88.79 & 100.00 & 96.26 \\
6 & 2 & 2 & 99.84 & 84.13 & 99.29 & 94.42 \\
6 & 3 & 2 & 99.84 & 85.08 & 99.92 & 94.95 \\
6 & 4 & 2 & 99.84 & 85.40 & 99.13 & 94.79 \\
6 & 5 & 2 & 99.84 & 87.14 & 100.00 & 95.66 \\
8 & 1 & 1 & 100.00 & 89.70 & 99.55 & 96.41 \\
8 & 2 & 1 & 100.00 & 87.58 & 99.70 & 95.76 \\
8 & 2 & 2 & 99.84 & 82.70 & 99.76 & 94.10 \\
8 & 3 & 2 & 99.84 & 81.11 & 99.84 & 93.60 \\
8 & 4 & 1 & 100.00 & 71.11 & 99.81 & 90.31 \\
8 & 4 & 2 & 100.00 & 80.30 & 100.00 & 93.43 \\
8 & 5 & 1 & 100.00 & 77.58 & 99.85 & 92.47 \\
8 & 5 & 2 & 99.84 & 80.32 & 99.52 & 93.23 \\
10 & 2 & 2 & 99.84 & 79.84 & 100.00 & 93.23 \\
10 & 3 & 2 & 99.84 & 81.27 & 99.84 & 93.65 \\
10 & 4 & 2 & 99.84 & 81.43 & 99.52 & 93.60 \\
10 & 5 & 2 & 99.84 & 80.48 & 99.68 & 93.33 \\
12 & 1 & 1 & 100.00 & 80.91 & 100.00 & 93.64 \\
12 & 2 & 1 & 100.00 & 79.39 & 99.70 & 93.03 \\
12 & 2 & 2 & 100.00 & 83.64 & 100.00 & 94.55 \\
12 & 3 & 1 & 100.00 & 75.45 & 99.85 & 91.77 \\
12 & 3 & 2 & 100.00 & 85.15 & 99.85 & 95.00 \\
12 & 4 & 1 & 100.00 & 83.33 & 99.70 & 94.34 \\
12 & 4 & 2 & 100.00 & 82.73 & 99.55 & 94.09 \\
12 & 5 & 1 & 100.00 & 77.58 & 99.85 & 92.47 \\
12 & 5 & 2 & 100.00 & 79.39 & 100.00 & 93.13 \\
16 & 1 & 1 & 100.00 & 77.27 & 99.70 & 92.32 \\
16 & 2 & 1 & 100.00 & 89.39 & 99.85 & 96.41 \\
16 & 2 & 2 & 100.00 & 83.39 & 100.00 & 94.41 \\
16 & 3 & 1 & 100.00 & 85.15 & 99.85 & 95.00 \\
16 & 3 & 2 & 99.84 & 90.00 & 100.00 & 96.61 \\
%\rowcolor[HTML]{C0C0C0} 
16 & 4 & 1 & \cellcolor[HTML]{C0C0C0}\textbf{100.00} & \cellcolor[HTML]{C0C0C0}\textbf{90.30} & \cellcolor[HTML]{C0C0C0}\textbf{100.00} & \cellcolor[HTML]{C0C0C0}\textbf{96.77} \\
16 & 4 & 2 & 99.84 & 86.98 & 100.00 & 95.61 \\
16 & 5 & 1 & 100.00 & 83.64 & 100.00 & 94.55 \\
16 & 5 & 2 & 100.00 & 87.88 & 99.85 & 95.91 \\
18 & 2 & 2 & 99.84 & 88.25 & 99.92 & 96.01 \\
18 & 3 & 2 & 99.84 & 85.87 & 99.92 & 95.21 \\
18 & 4 & 2 & 99.84 & 86.03 & 100.00 & 95.29 \\
18 & 5 & 2 & 99.84 & 86.03 & 99.92 & 95.26 \\
20 & 1 & 1 & 99.91 & 74.82 & 99.86 & 91.53 \\
20 & 2 & 1 & 100.00 & 80.61 & 100.00 & 93.54 \\
20 & 2 & 2 & 100.00 & 86.06 & 100.00 & 95.35 \\
20 & 3 & 1 & 100.00 & 83.94 & 99.85 & 94.60 \\
20 & 3 & 2 & 99.84 & 83.17 & 99.76 & 94.26 \\
20 & 4 & 1 & 100.00 & 70.30 & 99.55 & 89.95 \\
20 & 4 & 2 & 99.84 & 82.38 & 99.92 & 94.05 \\
20 & 5 & 1 & 100.00 & 68.48 & 99.55 & 89.34 \\
20 & 5 & 2 & 100.00 & 83.94 & 99.70 & 94.55 \\
22 & 2 & 2 & 99.84 & 84.29 & 99.92 & 94.68 \\
22 & 3 & 2 & 99.84 & 82.22 & 100.00 & 94.02 \\
22 & 4 & 2 & 99.84 & 83.49 & 99.76 & 94.37 \\
22 & 5 & 2 & 99.84 & 80.63 & 99.84 & 93.44 \\
24 & 2 & 2 & 99.84 & 86.19 & 99.76 & 95.26 \\
24 & 3 & 2 & 99.84 & 84.44 & 99.92 & 94.74 \\
24 & 4 & 2 & 99.84 & 82.38 & 99.84 & 94.02 \\
24 & 5 & 2 & 99.84 & 80.63 & 99.76 & 93.41 \\
26 & 2 & 2 & 99.84 & 82.70 & 99.92 & 94.15 \\
26 & 3 & 2 & 99.84 & 82.70 & 99.84 & 94.13 \\
26 & 4 & 2 & 99.84 & 81.43 & 99.92 & 93.73 \\
26 & 5 & 2 & 99.84 & 82.54 & 100.00 & 94.13 \\
28 & 2 & 2 & 99.84 & 81.90 & 99.84 & 93.86 \\
28 & 3 & 2 & 100.00 & 80.00 & 100.00 & 93.33 \\
28 & 4 & 2 & 100.00 & 79.70 & 98.79 & 92.83 \\
28 & 5 & 2 & 99.84 & 79.84 & 99.29 & 92.99 \\
30 & 2 & 2 & 99.84 & 80.48 & 100.00 & 93.44 \\
30 & 3 & 2 & 99.84 & 79.37 & 100.00 & 93.07 \\
30 & 4 & 2 & 99.84 & 76.83 & 100.00 & 92.22 \\
30 & 5 & 2 & 99.84 & 77.78 & 99.92 & 92.51 
\end{longtable}

%%%%%%%%%%%%%%%%%%%%%%%%%%%%%%%%%%%%%%%%%%%%%%%%%%%%%%%%%%%%

\end{document}